%% file: sample-sigconf.tex
\documentclass[sigconf]{acmart}

\usepackage{booktabs} 
\usepackage{subfigure}
\usepackage{todonotes}

\acmConference[]{March 2018}{Imperial College London}{UK}

\begin{document}
\title{Generalised Structural CNNs (SCNNs) for time series data with arbitrary graph topology}

\author{Thomas Teh}
\orcid{1234-5678-9012}
\affiliation{%
  \institution{Brain \& Behaviour Lab, Imperial College London}
  \streetaddress{Department of Computing}
  \city{London}
  \state{UK}
  \postcode{SW7 2BP}
}
\email{gim.teh16@imperial.ac.uk}

\author{Chaiyawan Auepanwiriyakul}
\affiliation{%
  \institution{Brain \& Behaviour Lab,  Imperial College London}
  \streetaddress{Department of Bioengineering}
  \city{London}
  \state{UK}
  \postcode{SW7 2BP}
}
\email{chaiyawan.aupenwiriyakul16@imperial.ac.uk}

\author{John A. Harston}
\affiliation{%
  \institution{Brain \& Behaviour Lab, Imperial College London}
  \streetaddress{Department of Bioengineering}
  \city{London}
  \country{UK}
}
\email{j.harston17@imperial.ac.uk}

\author{A. Aldo Faisal}
\affiliation{%
  \institution{Brain \& Behaviour Lab, Imperial College London}
  \streetaddress{Department of Bioengineering}
  \city{London}
  \country{UK}
}
\email{a.faisal@imperial.ac.uk}

\renewcommand{\shortauthors}{Teh et al.}

\begin{abstract}
Deep Learning methods, specifically convolutional neural networks (CNNs), have seen a lot of success in the domain of image-based data, where the data offers a clearly structured topology in the regular lattice of pixels. This 4-neighbourhood topological simplicity makes the application of convolutional masks straightforward for time series data, such as video applications, but many high-dimensional time series data are not organised in regular lattices, and instead values may have adjacency relationships with non-trivial topologies, such as small-world networks or trees. In our application case, human kinematics, it is currently unclear how to generalise convolutional kernels in a principled manner. Therefore we define and implement here a framework for general graph-structured CNNs for time series analysis. Our algorithm automatically builds convolutional layers using the specified adjacency matrix of the data dimensions and convolutional masks that scale with the hop distance. In the limit of a lattice-topology our method produces the well-known image convolutional masks. We test our method first on synthetic data of arbitrarily-connected graphs and human hand motion capture data, where the hand is represented by a tree capturing the mechanical dependencies of the joints. We are able to demonstrate, amongst other things, that inclusion of the graph structure of the data dimensions improves model prediction significantly, when compared against a benchmark CNN model with only time convolution layers.
\end{abstract}

%
%



\maketitle

\input{samplethomasbody-conf}

\clearpage

\bibliographystyle{ACM-Reference-Format}
\bibliography{sample-bibliography}

\end{document}

%% file: samplethomasbody-conf.tex
\section{Introduction}

\indent The success of deep learning, specifically convolutional neural networks (CNNs), in computer vision \citep{krizhevsky2012imagenet} has spurred applications of deep learning methods to domains such as natural language processing \citep{gehring2017convolutional,kalchbrenner2016neural,bradbury2016quasi}, speech recognition \citep{graves2013speech}, human activity recognition \citep{ordonez2016deep,neverova2016learning,toshev2014deeppose} and weather forecasting \citep{xingjian2015convolutional}. By design, CNNs share parameters across the input features and have sparse connections between layers, making them effective and efficient models for exploiting the local stationarity and lattice topology of pixels in an image. Similarly, recurrent neural networks (RNNs) and sliding windows, which extract features from temporal data by reusing the model parameters across different time steps, implicitly assume a stationary distribution within the input.

\indent In the realm of human activity modeling, the modelling processes can be broadly categorized into two categories - activity recognition and activity pattern detection. Activity recognition focuses on detection and classification of predetermined activities \citep{ordonez2016deep,yang2015deep} or surveillance technology \citep{neverova2016learning,boominathan2016crowdnet}, while deep learning techniques have been used in modelling human activity, with most studies focusing on activity recognition \citep{ordonez2016deep,yang2015deep,du2015hierarchical,ji20133d,jain2016structural} and limited set performing unsupervised learning on human-activity data \citep{butepage2017deep,holden2016deep}.

\indent However, as with most CNN studies used outside of the Computer Vision domain, the use of CNNs in this case is improper, as CNNs are optimised for data with a lattice topology, such as the pixel array of an image. When used with data that doesn't conform to a lattice topology, CNN performance drops, as the convolutional function can not fully capture the correlations between neighbouring connected data nodes.

\indent The application of deep learning models such as CNNs to human kinematics data is thus not straightforward, as the structure of human motion capture data is subjected to the constraints of human anatomy \citep{lin2000modeling}. Unlike the regular lattice array of images, human motion capture data have a tree-like structure (each hand is attached to an arm, which is jointly attached to the trunk, etc.) \cite{lin2000modeling}. Moreover,  human kinematics data generally contains both spatial and temporal features, and it is important to be able to capture spatio-temporal correlations between the features. Most deep learning models are only adept at modeling spatial and temporal features separately \citep{simonyan2014two} or in a stage-wise manner \citep{yang2015deep,ordonez2016deep,sainath2015convolutional} - the applications of deep learning models to model spatio-temporal features simultaneously requires significant ingenuity in the design of either architecture or new artificial neuron units \citep{xingjian2015convolutional,du2015hierarchical,jain2016structural}.

We hereby demonstrate a novel CNN architecture that can deep learn time series data with an arbitrary graph structure. We combine work on adjacency matrices with traditional CNN and RNN architectures, to allow us to perform deep learning on human kinematics data. We present both a generative model and a predictive model, built with our novel architecture. We train and test several models including our own on in-house human kinematics data, and find that our Structural Convolutional Neural Networks (SCNNs) outperform time-based convolutional neural networks. We also find that within our Structural Convolution AutoEncoder (SCAE), the convolutional kernels learn to only represent ethologically relevant hand movements in a sparse manner.

\subsection{Modelling Human Kinematics Data}
\indent Human kinematics data is most often represented as graph-structured spatio-temporal data. This proves a major hurdle to accurate modelling - most techniques in this regard have historically fallen short, in having no spatial or temporal convolution, or through restricting rather than incorporating graph structure, resulting in suboptimal prediction performance. 

\indent One of the earliest and simplest approaches to modelling human kinematics is the 'sliding window' method \citep{ordonez2016deep,yang2015deep,ji20133d,butepage2017deep}, which outperforms all recurrent neural networks in short term prediction \citep{gers2001applying} for human activity recognition tasks \citep{yang2015deep,ordonez2016deep,roggen2010collecting,jain2016structural}. Whilst useful, this approach doesn't conserve the spatial correlations that exist within the input data. Another traditional approach to modelling human kinematics temporally involves building a single end-to-end architecture consisting of convolutional and recurrent layers in a stage-wise manner \cite{sainath2015convolutional, ordonez2016deep}. This approach, however, lacks the capability to work on an arbitrary graph structure, as it features a regular convolution function.   

To address the problems described above, several models have been proposed that feature a graph structured convolution. Li et al. \citep{li2015hierarchical} proposed 3 hand-crafted multi-stream bidirectional RNNs that models each part of the body separately. Even though these models have hierarchical feature extractions that allow them to achieve better classification accuracy, their fusing layers do not account for the correlation of data prior to passing into the bidirectional layers. In addition, 2 out of 3 models fail to account for the structure of and the correlation between the spatial features. 

Another approach is the tree-based CNN, originally introduced in the natural language processing domain \citep{mou2015discriminative,mou2015natural,mou2016recognizing,collins2002convolution}. In this model, the input to the neural network needs to be organized hierarchically in a tree graph which allows for hierarchical feature extraction. However, this model also restricts the data structure into that of a tree, not allowing for arbitrarily-defined structure. A structural recurrent neural network was proposed by \citep{jain2016structural} in order to model spatio-temporal data with such arbitrary graph structure. Whilst this approach might generate human-like motion, this model is prone to the long-term dependencies problem common to all RNN models.

\section{Methodology}
\subsection{Data Acquisition \& Preprocessing}
\begin{figure}
	\begin{center}
    \subfigure[Joint locations]{
        \label{fig:handsens}
        \includegraphics[width = 0.75\hsize]{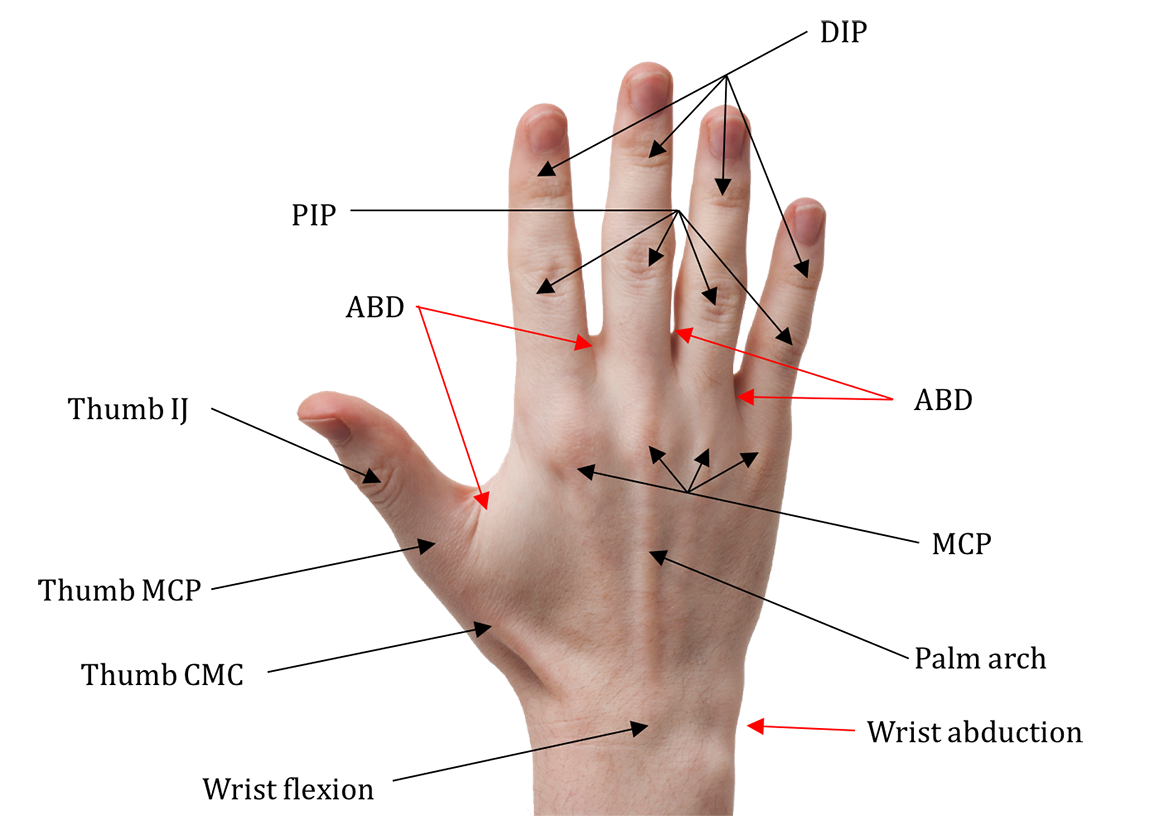}} 
    \subfigure[Graphical model of an image.]{
    	\label{fig:imagegraphical}
    	\includegraphics[width = 0.45\hsize]{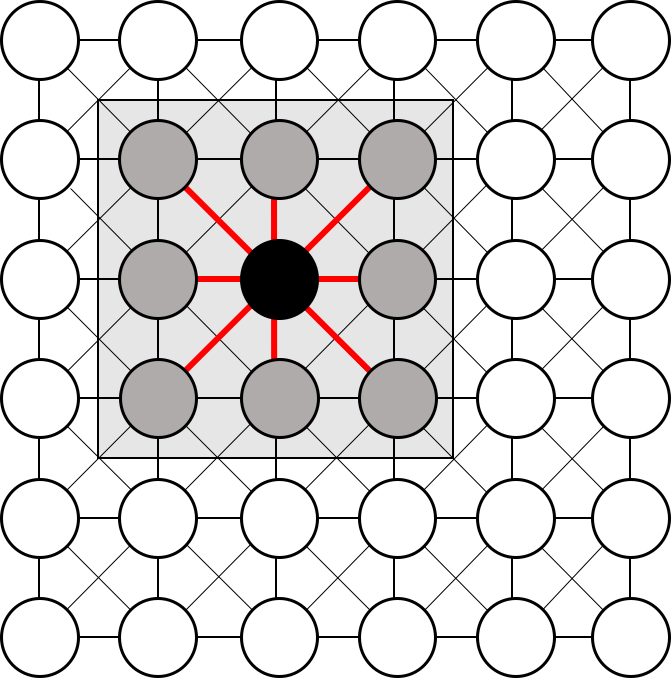}}
	\subfigure[Graphical model of a human hand.]{
      	\label{fig:handtree}
      	\includegraphics[width = 0.45\hsize]{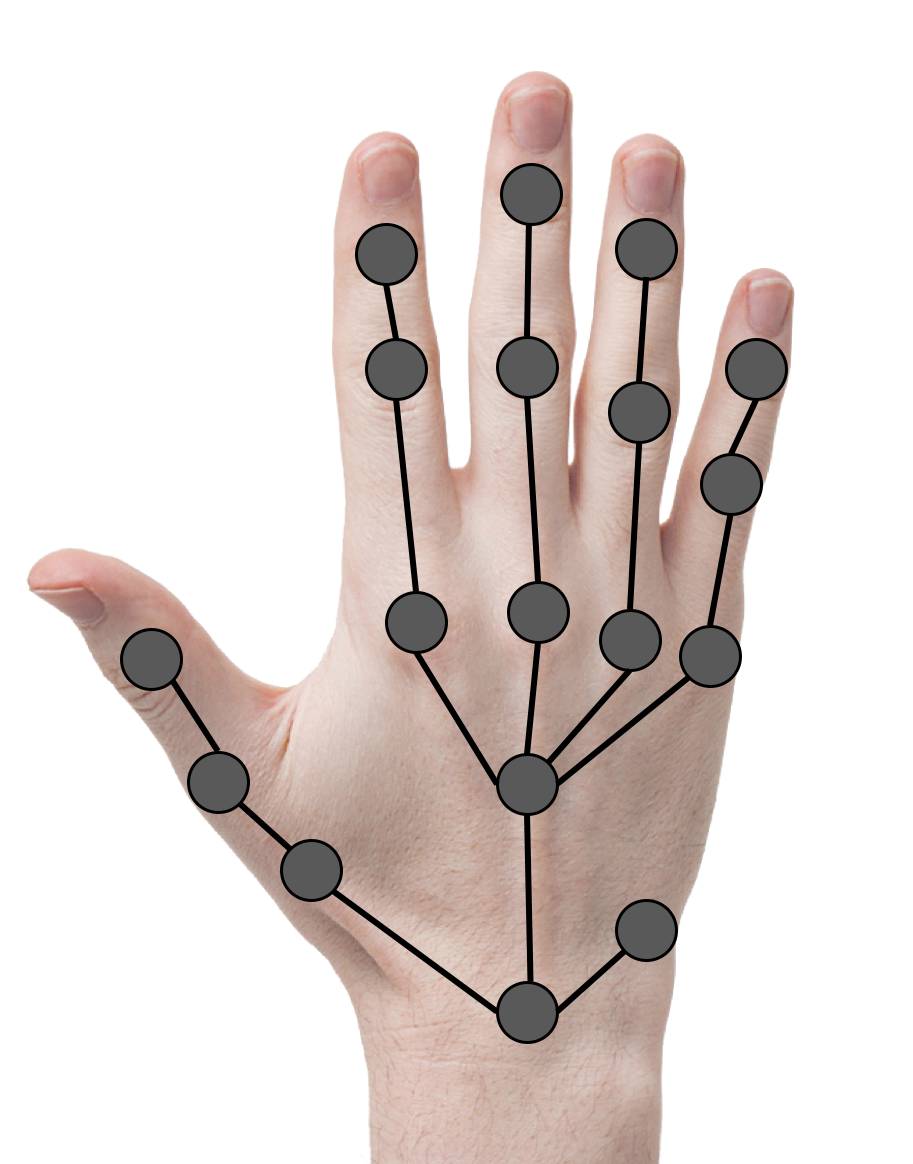}}
	\subfigure[Hand adjacency matrix]{
        \label{fig:adjmathand}
        \includegraphics[width = 0.75\hsize]{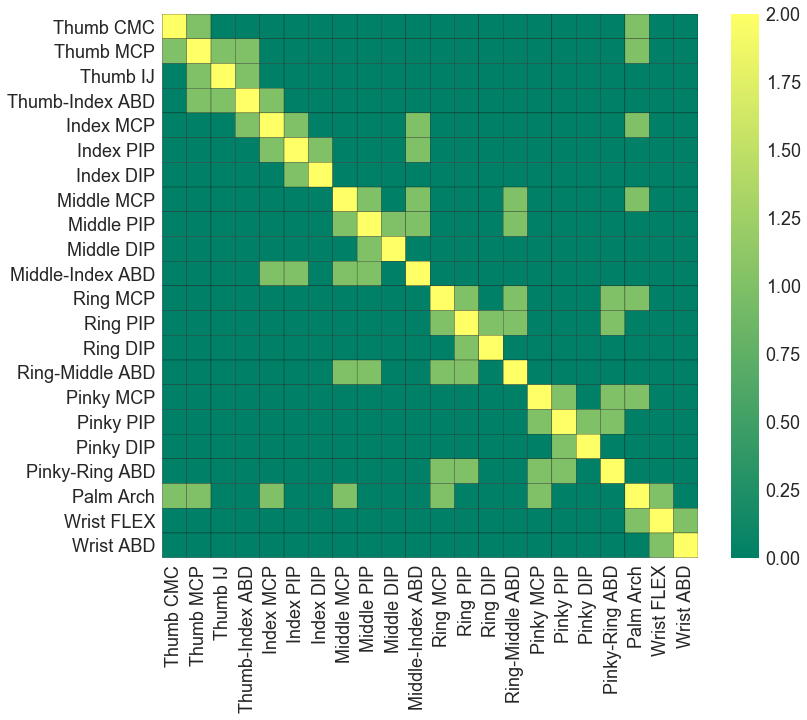}}
    \caption{(a) Locations of the 22 sensors embedded in CyberGlove, used to measure the angle of the joints (15 sensors),  the abduction between fingers (4), wrist flexion, wrist abduction and palm arch (1 each). (b,c) The features for an image are arranged on a grid, for a given  node on the lattice (black node). It has high correlation with its neighbors (grey nodes), whereas the features of the hand motion data set can be arranged according to the anatomical structure of the hand. (d) Adjacency matrix for the hand.}
    \label{fig:hand} 
	\end{center}
\end{figure}

We captured natural hand movements during daily life activities in our research group (following \cite{belic2015decoding}).The glove was calibrated against optical motion tracking methods using \cite{vicente2013calibration}.  All subjects gave written consent and the experimental procedure was approved by a local ethics committee. 
Subjects (N=10) wore a right-hand CyberGlove (CyberGlove Systems LLC, San Jose, CA, U.S.A.).  The glove measures joint abduction in 22 hand joints using stretch sensors embedded in the material with a spatial resolution of <1 degree (see Fig.~\ref{fig:hand} for joints tracked) and a sampling rate of 90 Hz. We recorded multiple hours of data per subject, yielding over 5 million samples.

\subsection{Structural Convolutional Neural Networks}

To both capture the spatio-temporal correlation from within the graph, and to work with any arbitrary graph structure, we propose a novel deep learning architecture, the Structural Convolutional Neural Network (SCNN). Our network design builds on several studies \citep{du2015hierarchical,butepage2017deep,mou2015discriminative,bruna2013spectral,henaff2015deep,mou2015natural,mou2016recognizing,jain2016structural,niepert2016learning,bronstein16geometric} that attempt to embed graph structure into the neural network itself. In contrast to these previous methods, our neural network architectures defines and uses  specialized convolutional kernel with an arbitrarily definable adjacency matrix. This enables us to embed prior knowledge for example in the form of physically known neighbourhood relationships between sensors.

To help explain our network architecture, we first defined the following for a graph with $F$ number of nodes and the adjacency matrix $\vec{A} \in \mathbb{R}^{F\times F}$:
\begin{align}
	\vec{y}^{\ell-1}	&= \begin{bmatrix}
	\vec{y}_1^{\ell-1}, \ldots, \vec{y}_F^{\ell-1}
    \end{bmatrix}^\top,  &\textit{Previous layer's output}\\
	\vec{y}^\ell  &= \begin{bmatrix}
	\vec{y}_1^\ell, \ldots, \vec{y}_F^\ell
	\end{bmatrix}^\top,   &\textit{Current layer's output}\\
    \vec{W}^\ell & = \begin{bmatrix}
    \vec{W}^\ell_1, \ldots, \vec{W}^\ell_F
    \end{bmatrix}^\top, &\textit{Current layer's weights}\\
	\vec{b}^\ell & = \begin{bmatrix}
    \vec{b}_1, \ldots, \vec{b}_F
    \end{bmatrix}^\top, &\textit{Current layer's biases}
\end{align}
where
\begin{align*}
\vec{y}^{\ell-1} 	& \in \mathbb{R}^{T\times F \times N },\\
\vec{y}^{\ell} 		& \in \mathbb{R}^{ (T-(t-1))\times F \times M},\\
\vec{W}^{\ell} 		& \in \mathbb{R}^{F \times t \times F \times N \times M},\\
\vec{b}^{\ell} 		& \in \mathbb{R}^{F \times M},\\
\vec{y}^{\ell-1}_i 	& \in \mathbb{R}^{T \times 1 \times N}, \forall i =1, \ldots, F,\\ 
\vec{y}^{\ell}_i 	& \in \mathbb{R}^{(T-(t-1)) \times 1 \times N}, \forall i =1, \ldots, F,\\ 
\vec{W}^{\ell}_i 	& \in \mathbb{R}^{t \times F \times N \times M}, \forall i =1, \ldots, F,\\
\vec{b}^{\ell}_i 	& \in \mathbb{R}^{1 \times M}, \forall i =1, \ldots, F.
\end{align*}

The kernel is made up of $F$ sub-kernels and each of the sub-kernels $i$, which corresponds to node $i$, has weights $W_i^\ell$ with the dimension of $t\times F\times N\times M$. The sub-kernels are slid across the temporal dimension of the input, producing an output of $(T-(t-1)) \times 1 \times M$ for each node $i$. The output is then passed through an activation function $g$ to produce:
\begin{align}
\vec{y}_i^\ell &= g\left(\vec{W}_i^\ell * \vec{y}^{\ell-1} + \vec{b}_i^\ell\right)\\
\vec{W}_i^\ell &=\begin{bmatrix} \vec{w}_{i1}^\ell \\  \vdots \\  \vec{w}_{iF}^\ell\end{bmatrix}
\end{align}
where $*$ is the convolution operation,
\begin{align}
	\vec{W}_i^{\ell} * \vec{y}^{\ell-1} & = \sum_{j=1}^F \vec{w}_{ij}^\ell * \vec{y}_j^{\ell-1} 
\end{align}
and $\vec{w}_{ij}^\ell$ is the sub-kernel weights for the $i$ node with its $j$ neighbor,
\begin{align}
	\vec{w}_{ij}^\ell \in
	\begin{cases}
		\mathbb{R}^{t \times 1 \times N \times M},&\textit{if $\vec{A}_{ij}\neq 0$},\\
		\mathbf{0}^{t \times 1 \times N \times M},&\textit{if $\vec{A}_{ij}=0$}.
    \end{cases}
\end{align}

where $\vec{w}_{ij}^\ell$ represents the sub-kernel and $\vec{A} \in \mathbb{R}^{F\times F}$ represents the adjacency matrix. Thus, if applied to the following graph:

\begin{figure}[H]
	\begin{center}
		\includegraphics[width = .65 \hsize]{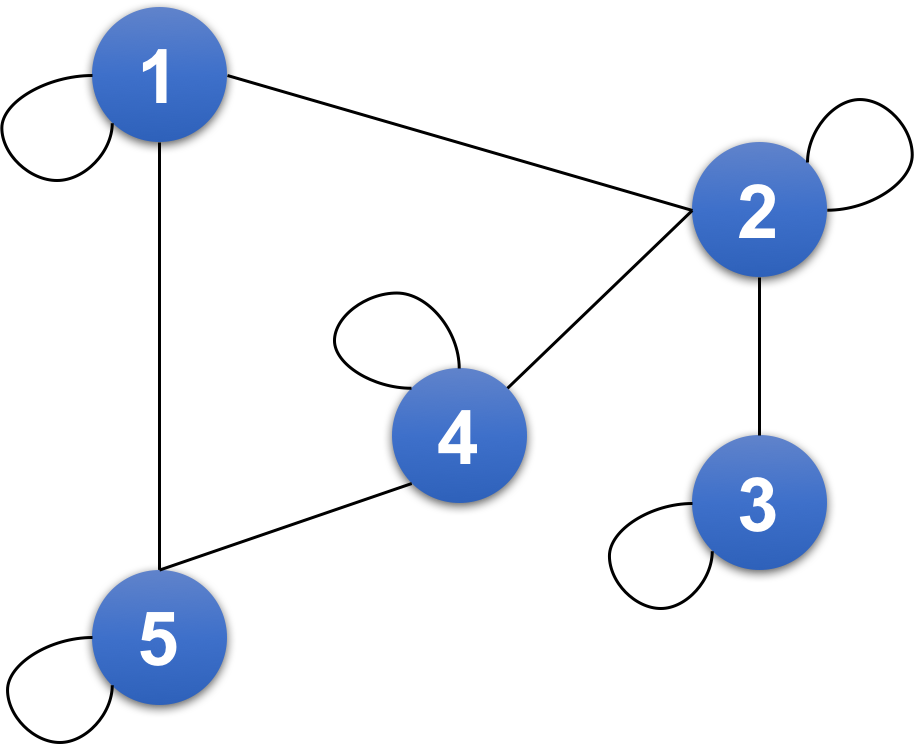} 
		\caption{Example of dependency graph to be modeled by our structural convolutional neural networks.}
		\label{fig:SampleGraph} 
	\end{center}
\end{figure}   

The graph can be represented by the adjacency matrix, $\vec{A}$: 
\begin{align}
	\vec{A} & = \begin{bmatrix}
    2	&	1	&	0	&	0	&	1\\
    1	&	2	&	1	&	1	&	0\\
    0	&	1	&	2	&	0	&	0\\
    0	&	1	&	0	&	2	&	1\\
    1	&	0	&	0	&	1	&	2
    \end{bmatrix}.
\end{align}

The kernel weights $\vec{W}^\ell$ consists of the sub-kernels with the corresponding weights, $\vec{W}^\ell_i$ below:
\begin{align}
	\vec{W}^\ell_1 &= \begin{bmatrix}
    \vec{w}_{11}^\ell	& 	\vec{w}_{12}^\ell	& 	\vec{0}		&	\vec{0}		&	\vec{w}_{15}^\ell \end{bmatrix}^\top\\
    \vec{W}^\ell_2 &= \begin{bmatrix}
    \vec{w}_{21}^\ell	& 	\vec{w}_{22}^\ell	& 	\vec{w}_{23}^\ell		&	\vec{w}_{24}^\ell		&	\vec{0} \end{bmatrix}^\top\\
    \vec{W}^\ell_3 &= \begin{bmatrix}
    \vec{0}	& 	\vec{w}_{32}^\ell	& 	\vec{w}_{33}^\ell		&	\vec{0}		&	\vec{0} \end{bmatrix}^\top\\
    \vec{W}^\ell_4 &= \begin{bmatrix}
    \vec{0}	& 	\vec{w}_{42}^\ell	& 	\vec{0}		&	\vec{w}_{44}^\ell		&	\vec{w}_{45}^\ell \end{bmatrix}^\top\\
    \vec{W}^\ell_5 &= \begin{bmatrix}
    \vec{w}_{51}^\ell	& 	\vec{0}	& 	\vec{0}		&	\vec{w}_{54}^\ell		&	\vec{w}_{55}^\ell \end{bmatrix}^\top.
\end{align}

\begin{figure}[H]
	\centering 
	\subfigure[Mechanics of the sub-kernels on the input layer. The sub-kernels are represented by the dotted rounded rectangles.]{\label{fig:subkernels}\includegraphics[width = 0.55\hsize]{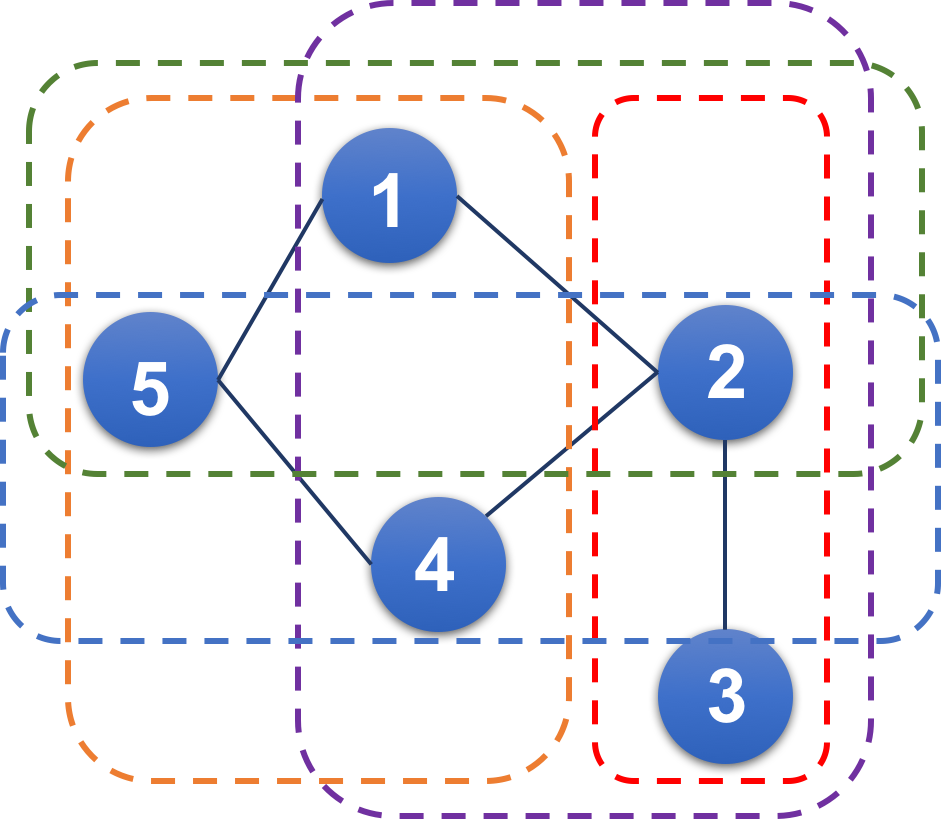}}
    \hspace{15mm} 
\subfigure[Structural convolutional layer. The number in the nodes represents the nodes in the input that are convolved.]{\label{fig:subkernelconvolution}\includegraphics[width = 0.55\hsize]{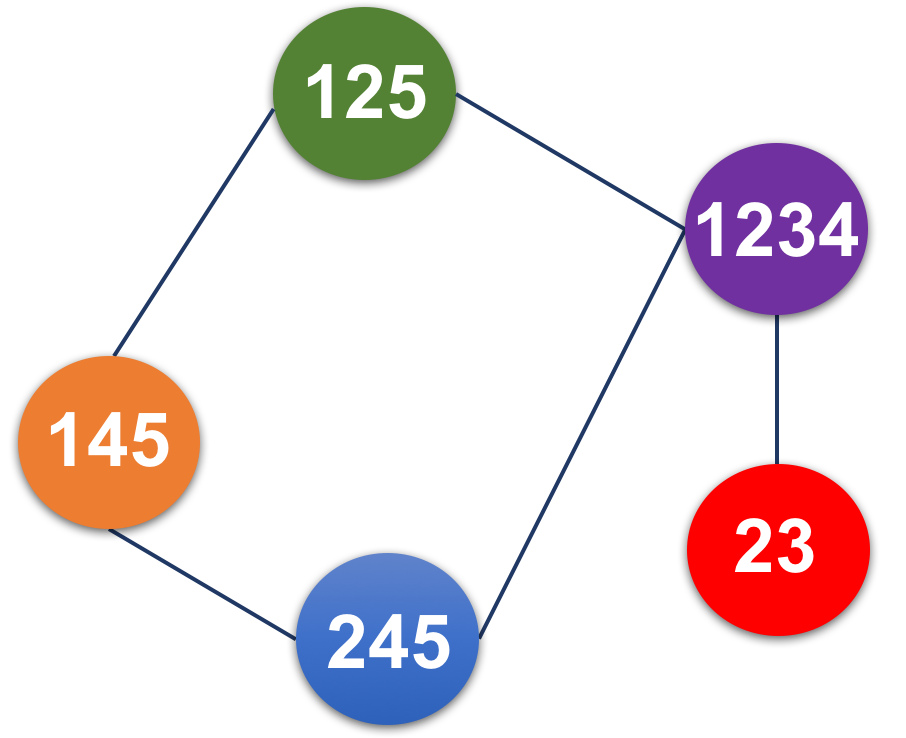}}
\caption{The sub-kernels convolve only specific nodes in the input layer to produce the corresponding nodes in the convolutional output layer. For example, the sub-kernel that encompasses the input nodes 1, 2, 3 and 4 maps those input nodes to the purple node in the convolutional layer. The structure of the graph remains intact after the convolution operation. The recurrent edges are omitted for brevity.}
\end{figure}

Figure \ref{fig:subkernels} and \ref{fig:subkernelconvolution}  shows the workings of the structural convolution for the graph in Figure \ref{fig:SampleGraph}, for input with a single channel and a single kernel. Each of the sub-kernels will only take some of the nodes of the graph for the convolution operation. Furthermore, each of the sub-kernels is distinct to the input nodes. For example, in Figure \ref{fig:subkernels}, the sub-kernel for node 2 (in purple) will take all the neighbors of node 2 that are 1 path length away for the convolution operation. The output for the convolution operation is then mapped to the corresponding node on the convolution layer.

Additionally, the use of an adjacency matrix allows an arbitrary graph structure to embed itself into the core convolution function and thus preserve any spatial correlations the data might possess prior to and after the convolution function. Furthermore, in contrast to the hard limit placed on the number of possible node connections in the previous study \citep{niepert2016learning}, our method allows all nodes that are reachable within a predetermined path to be covered by one convolution function and, as a result, allows for more efficient construction of CNNs for data that have large graph structures.

\subsection{Neural Network Structure} 
We present two novel neural network architectures that leverage our graph-structural approach: 1. A structural convolutional autoencoder, and 2. A structural convolutional neural network, in order to test our architecture in both supervised and unsupervised learning environments.

\begin{figure}[ht]
	\begin{center}
        \includegraphics[width = 1.0\hsize]{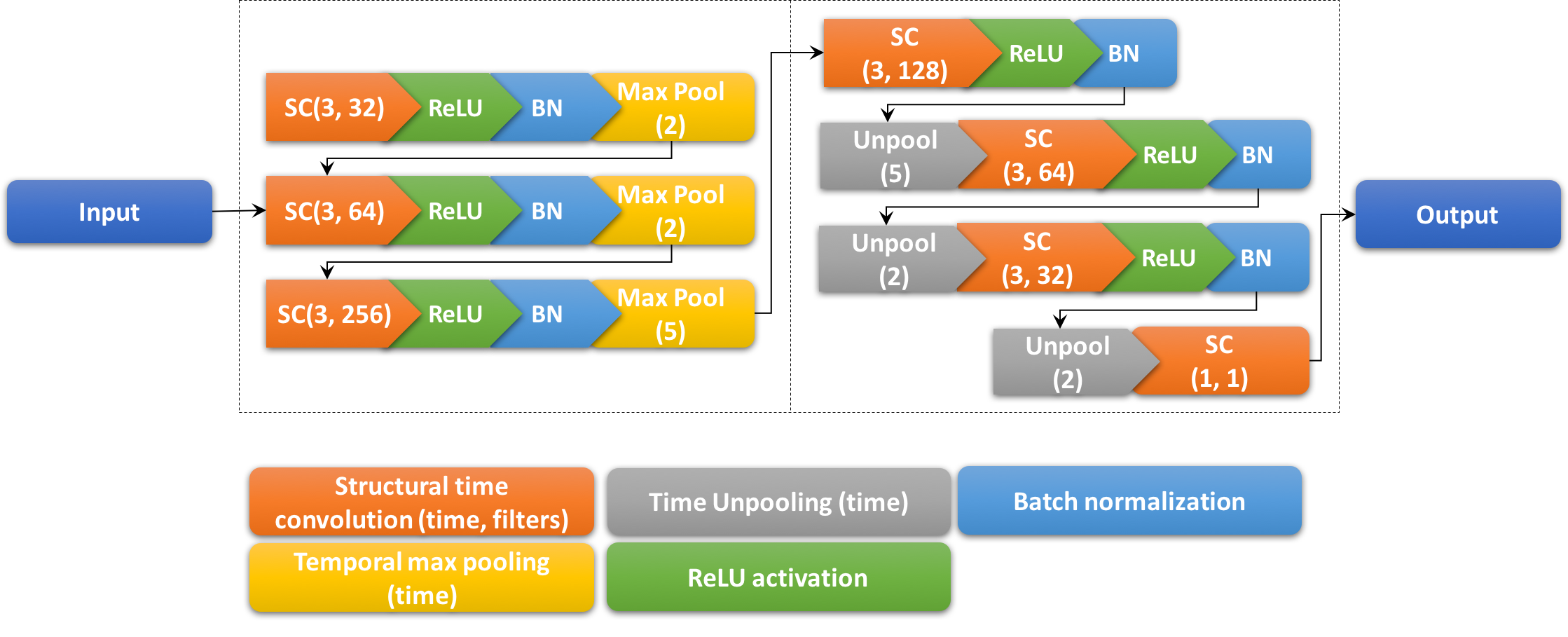}
		\caption{Structural convolutional autoencoder (SCAE). Similarly to the SCNN, the convolution operation is performed on both the spatial and temporal dimensions, with differences in the deconvolution process, with respect to the temporal unpooling and fully-connected layers.}
		\label{fig:scae} 
	\end{center}
\end{figure}

\paragraph{Structural Convolutional AutoEncoder (SCAE)} For unsupervised learning, this study implements the structural convolutional autoencoder as shown in Figure \ref{fig:scae}. The model is trained initially without any regularization until weights are relatively stable. Thereafter, we impose the $L1$ regularization penalty to fine-tune the weights further.

\begin{figure}[ht]
	\begin{center}
        \includegraphics[width = 1.0\hsize]{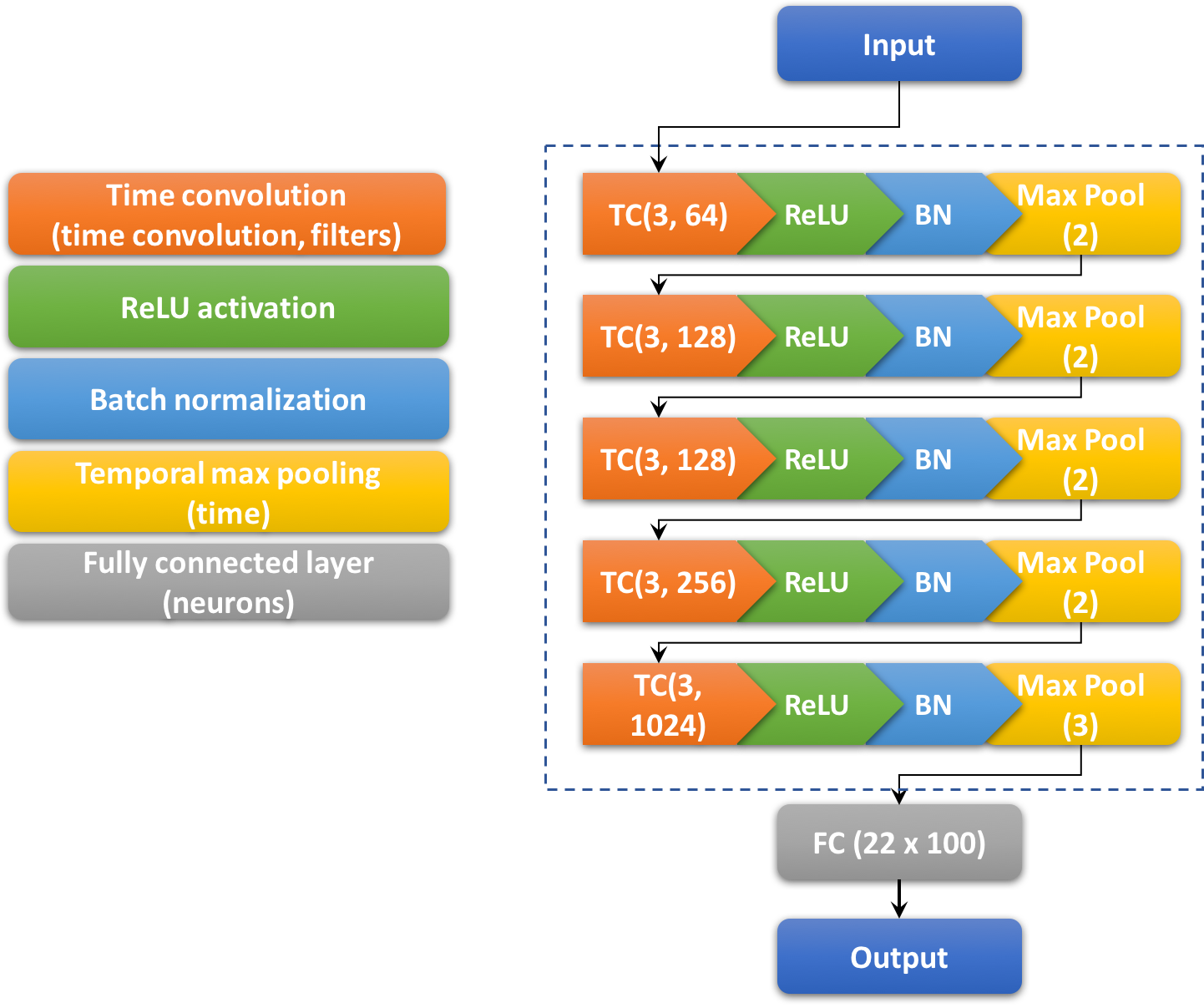}
		\caption{Time convolutional neural networks (TCNNs). Notations for the blocks are similar to Figure \ref{fig:scnn}, with the exception of the convolutional layer. The convolutional layer in this model, TC(time steps, filters) only convolves the input temporally.}
		\label{fig:tcnn} 
	\end{center}
\end{figure}

\begin{figure}
	\begin{center}
        \includegraphics[width = 1.0\hsize]{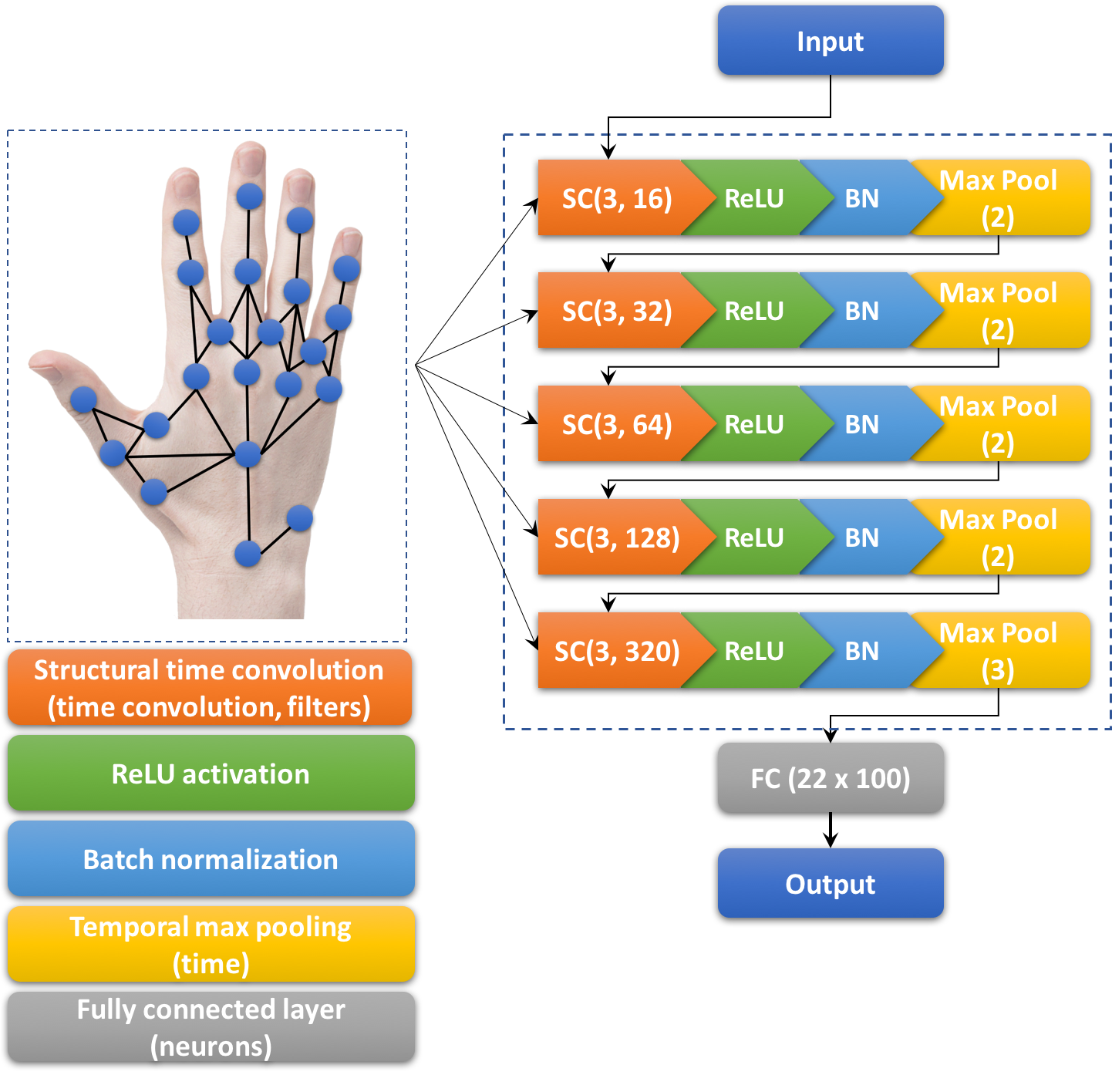}
		\caption{Structural convolutional neural networks (SCNN). The convolution operation is done on both spatial and temporal dimensions. SC(time steps, filters) denotes the number of time steps to convolve and the number of feature maps produced from the structural convolution. ReLU and BN refer to the ReLU activation layer and batch normalization respectively. MaxPool(time steps) denotes the temporal max pooling and FC(number of neurons) denotes the fully connected layer.}
		\label{fig:scnn} 
	\end{center}
\end{figure}

With the large number of parameters in the model, it is relatively easy to train such that it can reconstruct its input perfectly. However, such a model would not provide us with significant insight. As the $L1$ regularization penalty encourages sparsity within the model, by retraining the model with $L1$ regularization, we fine tune the kernel weights such that some weights will have zero-values. In essence, the model will subsequently prioritize features that are more representative of natural hand movement.
\paragraph{Structural Convolutional Neural Networks (SCNN)} For the prediction task, we implement two different models: structural convolutional neural networks (SCNN) as in Figure \ref{fig:scnn} and temporal convolutional neural networks (TCNN) as in Figure \ref{fig:tcnn}. The latter TCNN model is used as a baseline from which to benchmark the relative performance of the SCNN model. The input to both models is the subsample of the time series with a 500-step time window, while the output is the subsequent 100-step time shift.

\subsection{Neural Network Implementation}
We implemented our models using the Tensorflow package \citep{abadi2016tensorflow}. In addition, we also implement $L1$ regularization on the parameter optimization. The Xavier initialization \citep{glorot2010understanding} was implemented to randomly initialize the weights of the convolution kernels. All biases were initialized to 0.5. An ADAM optimization technique was implemented to carry out our parameter optimization \citep{kingma2014adam}. Since our system can train several neural networks in parallel, a batching process was implemented to split the training data into 32 batches. Additionally, to ensure fast convergence of the parameter values \citep{bengio2012practical} and to prevent overfitting, the training batches are not constructed by subsampling sequentially, but instead by randomly shuffling into batches. Lastly, batch normalization \citep{ioffe2015batch} was implemented to help reduce internal covariance shift, and thus ensure fast convergence of the trainable parameters.

\section{Results}
To train, test, and validate our neural network structure, the joint angle time series data were segregated into training, test and validation data with 55\%, 35\%, and 10\% proportions. For standardization purposes, the training dataset attributes such as mean and variance were subsequently used to standardize every dataset to have zero mean and unit variance. For the predictive model, the inverse transform was applied at the output layer of the SCNN. 
A 500-time-step sliding window (equivalent to 3.56 seconds at 140 Hz) was applied to separate each dataset into multiple time frames. Additionally, a 100-step time shift was also applied to create prediction windows.
\subsection{Kernels \& Activation Layers Visualisation}
\begin{figure}[ht]
	\begin{center}
        \includegraphics[width = 1.0\hsize]{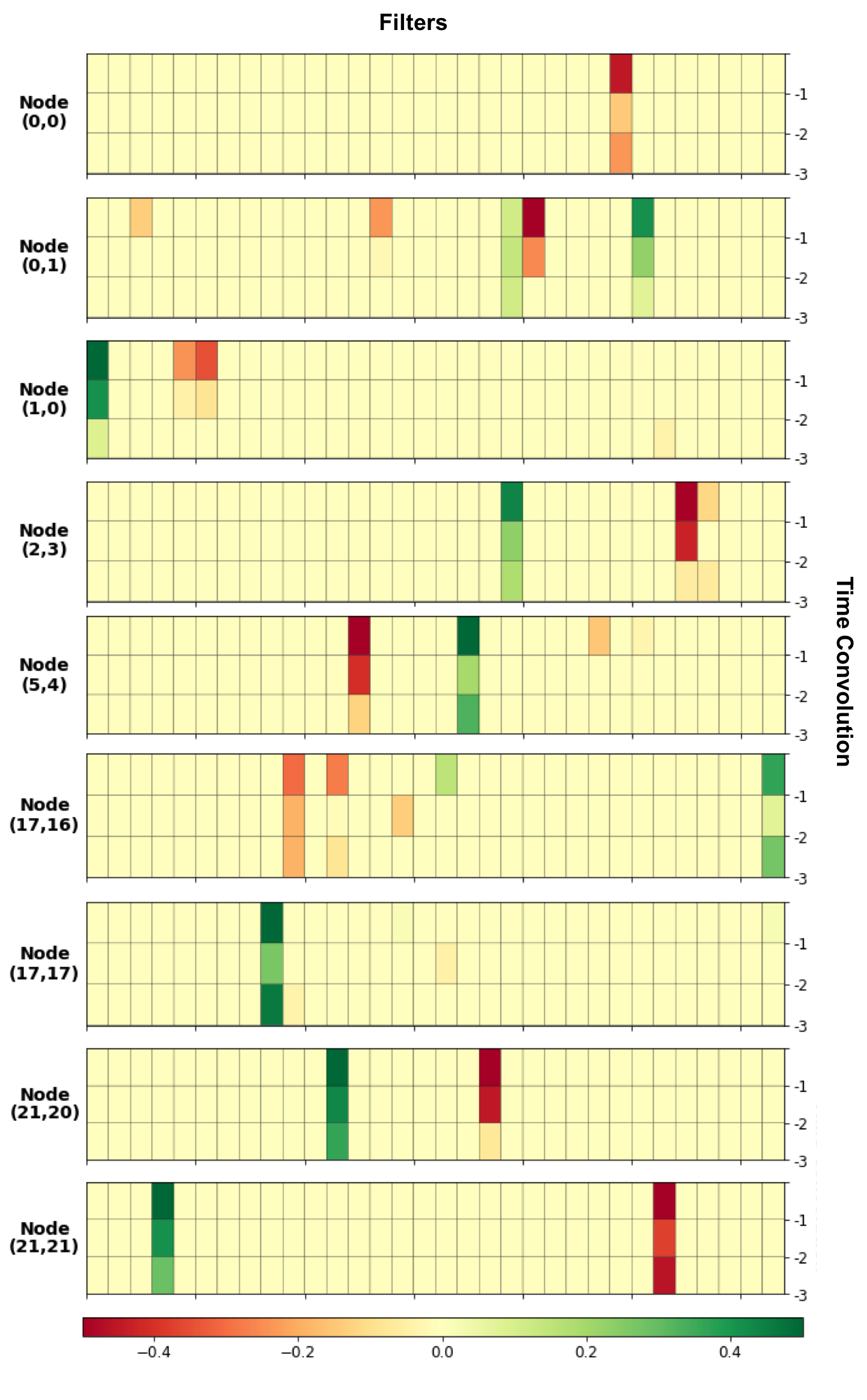}
		\caption{Visualization of the sub-kernel weights for the first layer. By imposing the $L1$ regularization, a large number weights are set to zero, and the non-zero weights are sufficient to reconstruct the input.}
		\label{fig:kernelvisualization} 
	\end{center}
\end{figure}

We selectively visualize the first layer kernels of the SCAE model to understand the representations of human motor dynamics. (Figure \ref{fig:kernelvisualization}). As the effects of the $L1$ regularization force most of the kernel weights to zero, the non-zero weights represent the most prominent motion features in the data. These also provide indications that we can further reduce the number of parameters in our model. 

For the kernels, the weights have the same sign across the time convolution, which implies the first layer captures the positions of the joints across time. Properties of the motion dynamics, such as velocity and acceleration, are likely to be captured in the deeper layers of the network.

\begin{figure}[t]
	\begin{center}
        \includegraphics[width = 0.95\hsize]{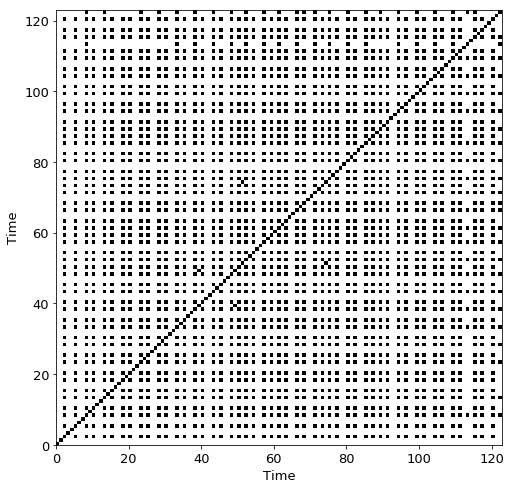}
		\caption{Recurrence plot for the activations of a specific feature map node in Layer 3. Threshold used to compute the recurrence is $1 \times 10^{-4}$.}
		\label{fig:recur} 
	\end{center}
\end{figure}

During training we notice a quasi-periodicity in the neural network activations. To confirm this phenomenon, we produced a recurrence plot of the activations (Figure \ref{fig:recur}), which shows a stereotypical quasi-periodic pattern, by using the following binary function:
\begin{align}
  R(t_i, t_j) &=  
  	\begin{cases}
  	1, & \parallel a(t_i) -a(t_j) \parallel_2 \leq \varepsilon\\
  	0, & \text{otherwise}
	\end{cases}
\end{align}
where $a(t)$ denotes the activation at time $t$ and $\varepsilon$ is the predetermined threshold. 

It can be observed that the SCAE is able to learn both the spatial and temporal structure in the data. The fact that only selected nodes have significant activations within the feature maps shows that the model learns the prominent spatial features. In addition, the quasi-periodicity of the activations indicates that the SCAE model also captures the temporal structure in the deeper layers of the model.

\subsection{Hand Movement Prediction}

The inclusion of the fully connected layer enables the model to predict with higher accuracy. However, unlike regular classification tasks where the number of classes tend to be small, for a regression task the fully-connected layer requires a much larger number of points to obtain accurate predictions for a longer prediction horizon.

\begin{figure*}[ht]
	\centering     
	\subfigure[Training data]{\label{fig:rmseconsotraining}\includegraphics[width = 0.49\hsize]{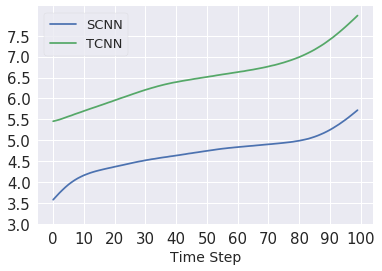}}
\subfigure[Test data]{\label{fig:rmseconsotest}\includegraphics[width = 0.50\hsize]{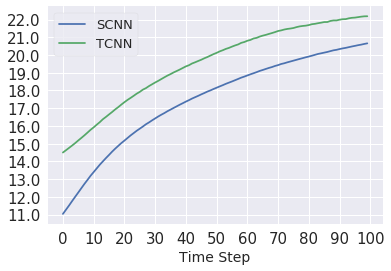}}
\caption{RMSE for TCNN and SCNN aggregated across all features. The TCNN is shown to perform consistently worse than the equivalent SCNN model.}
\label{fig:RMSE_Conso}
\end{figure*}

Our model extends the work in \citep{tehunpub2014bis} by including the graph structure of the features and predicting a fixed horizon instead of merely the next time step. We plot the aggregated RMSE in Figure \ref{fig:RMSE_Conso}.

The RMSE for the TCNN model is consistently higher than that for the SCNN model (Figure \ref{fig:RMSE_Conso}). Since the TCNN only incorporates the correlations between the spatial features at the fully-connected layer, the inclusion of the dependency graph of the spatial features in constructing the convolution layer is beneficial to the model's predictive power. 
    
Also from Figure \ref{fig:RMSE_Conso}, it is observed that the RMSE for both model worsens for predictions that are many time-steps ahead. The deterioration of the RMSE across time-steps is well within our expectations, as predictions that are significantly further ahead are naturally much less reliable.
    
Overall, both of our models can predict the movements of the data, even at a distant prediction horizon, however they fail to capture the magnitude of those movements. The SCNN outperforms the TCNN in terms of RMSE attained. These results validate the observations from \citep{butepage2017deep,du2015hierarchical} that the inclusion of the graph structure for human motion capture related tasks improves the prediction quality and allows us to lengthen the prediction horizon.


\section{Discussion}
Our main methodological contribution  is the introduction of the structural convolutional neural network, which allows efficient design of bespoke convolutional kernels via the specification of the dependency graph of the features. While this study focuses on the application of the model to human hand motion data, the proposed model can actually be applied to data with arbitrary topology. These special cases can be derived by specifying the adjacency matrices of the features in a specific manner. 

\paragraph{Prediction Model}
For prediction tasks, we compared two models: our structural convolutional neural networks (SCNNs) and  the well-known time convolutional neural networks (TCNNs). The difference between these two networks is that of inclusion of the adjacency matrix of the spatial features in the convolution masks. Based on the predictions we obtained for both models, we observed the following:
\begin{itemize}
	\itemsep0em
	\item  The SCNN model outperforms the TCNN model in terms of the RMSE and $R^2$ values. By embedding the topology of the spatial features of the data, the model is able include the local spatio-temporal interactions between the different joints in the early stages of the model.
	\item The improvement of the prediction for the SCNN stems from the inclusion of the graph structure allowing the neural network to extract more meaningful representations of movement dynamics, allowing for a higher accuracy across a longer prediction horizon.
\end{itemize}

Our approach allows us to design bespoke convolutional kernels using the adjacency matrix of the spatial features. We demonstrate here that our approach improves the prediction quality and extends the prediction horizon significantly.
This efficiency comes at a price: Unlike the structural RNN by \citep{jain2016structural} and the graph CNN by \citep{niepert2016learning}, our current approach does not support directed edges or edges features and is limited to undirected graphs. Thus, a natural extension of this study would be the inclusion of RNNs to the SCNNs by constructing a structural convolutional recurrent neural network similar to the convolutional LSTM in \citep{xingjian2015convolutional}, as a combined architecture may be better able to capture long-term spatio-temporal correlations. Beyond the inclusion of RNNs our convolutional kernel construction method can be improved by unsupervised graph structure estimation from the data.

We applied this approach to  the graph structure of the human body kinematics time series, and show that we outperform conventional time convolutional neural networks. Our approach allows the development of deep learning models trained on arbitrary graph structured data, be it medical data (e.g. fMRI-based brain network activity), economic data (e.g. airline travel numbers on the airport connectivity graph) or social data (e.g. social networks variables). The largest benefit for our method's flexibility is its scalability - the structural convolution requires a single adjacency matrix for all the spatial features.

\begin{acks}
This work has received funding from the European Union's Horizon 2020 research and innovation programme under grant eNHANCE (grant no 644000) -- www.enhance-motion.eu.

\end{acks}